\documentclass[conference]{IEEEtran}
\usepackage[cmex10]{amsmath}
\usepackage{amsmath}
\usepackage{amssymb}
\usepackage{bm}
\usepackage{algorithm}
\usepackage{algorithmic}
\usepackage{graphicx}
\graphicspath{{./fig/}}
\usepackage{multirow}

\ifCLASSOPTIONcompsoc
  \usepackage[caption=false,font=normalsize,labelfont=sf,textfont=sf]{subfig}
\else
  \usepackage[caption=false,font=footnotesize]{subfig}
\fi
\usepackage{url}
\begin{document}
%
\title{Learning Boltzmann Machine with EM-like Method}

\author{\IEEEauthorblockN{Jinmeng Song}
\IEEEauthorblockA{Department of Computer Science and Technology\\
Tsinghua University\\
Beijing 100084, China\\
Email: songjm12@mails.tsinghua.edu.cn}
\and
\IEEEauthorblockN{Chun Yuan}
\IEEEauthorblockA{Department of Computer Science and Technology\\
Tsinghua University\\
Beijing 100084, China\\
Email: yuanc@sz.tsinghua.edu.cn}}
\maketitle

\begin{abstract}
We propose an expectation-maximization-like(EM-like) method to train Boltzmann machine with unconstrained connectivity. It adopts Monte Calo approximation in the E-step, and replaces the intractable likelihood objective with efficiently computed objectives or directly approximates the gradient of likelihood objective in the M-step. The EM-like method is a modification of alternating minimization. We prove that EM-like method will be the exactly same with contrastive divergence in restricted Boltzmann machine if the M-step of this method adopts special approximation. We also propose a new measure to assess the performance of Boltzmann machine as generative models of data, and its computational complexity is O(Rmn). Finally, we demonstrate the performance of EM-like method using numerical experiments.
\end{abstract}

\section{Introduction}
Ackley et al.\cite{ackley1985learning} invented Boltzmann machine which could be regarded as stochastic version of Hopfiled network and Boltzmann machine is universal approximators of discrete distributions. So far, Boltzmann machine with unconstrained connectivity still can't be applied to any practical problem in machine learning or inference\cite{suzuki2013chaotic}. Although the original algorithm\cite{hinton1983optimal} training Boltzmann machine had a very simple form, the computational complexity is about $O(2^n)$. Different kinds of approximation methods have therefore been developed, including pseudo-likelihood estimation\cite{besag1975statistical,hyvarinen2006consistency}, contrastive divergence\cite{hinton2002training}, mean field theory\cite{Welling200319,yasuda2009approximate} and alternating minimization\cite{csisz1984information}.


Yasuda and Tanaka\cite{yasuda2009approximate} investigated approximate learning algorithm in Boltzmann machine using linear response estimate, the Bethe approximation and loopy belief propagation. In their paper, they noted that all of these algorithms would give poor results in Boltzmann machine with many loops.


When learning Boltzmann machine without hidden units, Hyv\"{a}rinen\cite{hyvarinen2006consistency} proved that pseudo-likelihood estimation provided a statistically consistent estimator. In pseudo-likelihood estimation, there isn't derivative of the log partition function which is a main barrier when one adopts maximum likelihood estimate to train Boltzmann machine.

Alternating minimization\cite{csisz1984information} is a popular approach to solve several optimization problems, such as channel capacity, rate-distortion functions and maximum likelihood estimate. Amari et al.\cite{amari1992information} studied the information geometry of Boltzmann machine. Based on the geometry of the neural manifold, they proposed a learning method which adopted alternating minimization and they proved its convergence. Based on alternating minimization, Byrne\cite{byrne1992alternating} presented another learning procedure for Boltzmann machine with unconstrained connectivity from different perspectives. The prerequisite of these two exact learning methods is that data distribution must be given directly, which is hard to achieve in practice.

Only restricted Boltzmann machine which is a two-layer machine without intralayer connections and Boltzmann machine without hidden units can be trained by contrastive divergence effectively. Empirically the contrastive divergence method can give a good solution, and some theoretical results about contrastive divergence have been reported\cite{carreira2005contrastive,yuille2006convergence}. It's modification, persistent contrastive divergence, was proposed by Tieleman\cite{tielemanPcd2008}. The major difference with standard contrastive divergence is that persistent ``fantasy particles'' are not reinitialized to data points after each weight update\cite{tieleman2009using}.

The expectation maximization (EM) algorithm is iterative procedure for obtaining maximum likelihood functions under models with hidden variables. When E-step is intractable, a modification of the EM algorithm where the expectation is computed numerically through Monte Carlo simulations can be used. This method was called Monte Carlo EM (MCEM) algorithm\cite{wei1990monte,lange1995gradient,fort2003convergence}.
The contribution in this paper includes:
\begin{enumerate}
\item An EM-like method proposed by us can train Boltzmann machine with any pattern of structure. It adopts Monte Calo approximation in the E-step and replaces the intractable likelihood objective with other objective or directly approximates the gradient of likelihood objective in the M-step.
\item We investigate the relationship of EM-like method with alternating minimization and contrastive divergence. Prove that the EM-like method is a modification of alternating minimization and contrastive divergence is only special case of EM-like method in restricted Boltzmann machine.
\item We propose a new measure to evaluate how the quality of Boltzmann machine meets the given data set, whose computational time is $O(Rmn)$.
\end{enumerate}

The content of this paper is organized as follows. Firstly, we describe the model of Boltzmann machine used in this paper. Secondly, we review how to adopt alternating minimization and MCEM to train Boltzmann machine. Then, EM-like method used in Boltzmann machine is described. After that, the detail of EM-like method and the relationship with other methods is studied. Thirdly, a new evaluating method is presented. Finally, the validity of EM-like method is verified using numerical experiments, and some concluding remarks are given in the end of this paper. 

\section{Boltzmann Machine Learning Problem}
Boltzmann machine discussed here is a network with stochastic binary units. Each state of Boltzmann machine is specified as $\bm{x} \in \{0, 1\}^{n}$. Boltzmann machine with $n$ units usually is divided into a set of visible units $\bm{v} \in \{0, 1\}^m$ and a set of hidden units $\bm{h} \in \{0,1\}^{n-m}$. It is customary that the first $m$ units are visible and we assume this henceforth. The entire state of Boltzmann machine $\bm{x}$ equals $\{\bm{v}, \bm{h}\}$.

The probability distribution of each state is defined as
\begin{align}
	\label{fun:distribution}
	p(\bm{v}, \bm{h}; \theta) = \frac{\exp\{-E(\bm{v}, \bm{h}; \theta)\}}{\mathcal{Z}(\theta)}
\end{align}
where $\theta = \{W, b\}$ is the parameter of Boltzmann machine, which contains a matrix $W \in R^{n\times n}$ and a vector $b \in R^{n}$; $\mathcal{Z}(\theta)$ is partition function, and its computational time is about $O(2^n)$. Thus, for any larger dimension $n$, direct numerical computation of $\mathcal{Z}(\theta)$ is computationally intensive.

Vector $b$ is called bias term. Matrix $W$ is pairwise interaction terms, which satisfies $w_{ii} = 0$ and $w_{ij} = w_{ji}$. For simplicity, we may use $p(\bm{x};\theta)$ and $E(\bm{x};\theta)$ to replace $p(\bm{v}, \bm{h};\theta)$ and $E(\bm{v}, \bm{h};\theta)$ respectively. Marginal distribution of $m$ visible units is determined from Boltzmann machine
\begin{equation}
	p(\bm{v};\theta) = \sum_{\bm{h}} p(\bm{v}, \bm{h};\theta)
\end{equation}

The learning problem in Boltzmann machine addressed here is to construct $n$-unit machine whose marginal distribution defined on visible units is closest to the given distribution $Q(V)$ defined on the set of states $\{0, 1\}^m$. Usually, the similarity between two distributions can be measured by Kullback-Leibler divergence
\begin{align}
	\label{fun:Kullback-Leibler divergence}
	D(P(Z)\|Q(Z)) = \sum_{z \in Z} p(z)\log\frac{p(z)}{q(z)}
\end{align}
where $P$ and $Q$ are arbitrary distribution which are defined on the same set of events $Z$. Using Kullback-Leibler divergence, learning problem can be restated as finding the optimal $n$-unit Boltzmann machine whose marginal distribution $P^*(V)$ is subject to
\begin{equation}
	\label{fun:loss_function}
	P^*(V) = \arg \min D(Q(V)\|P(V))
\end{equation}
where $P(V)$ is marginal distribution of a $n$-unit Boltzmann machine.

\subsection{Alternating Minimization in Boltzmann Machine}
We define the family of Boltzmann machine with $n$ units as
\begin{equation}
	\mathcal{B} = \left\{P(X) \in \mathcal{P}^n \left| p(\bm{x}; \theta) = \frac{\exp\{-E(\bm{x}; \theta)\}}{\mathcal{Z}(\theta)}\right. \right\}
\end{equation}
And we define the distribution family $\mathcal{E}_{Q(V)}$ where the marginal distribution of each element agrees with $Q(V)$.
That is:
\begin{align}
	\mathcal{E}_{Q(V)} = \left\{Q(V,H) \in \mathcal{P}^n\left|\sum_{\bm{h}} q(\bm{v}, \bm{h}) = q(\bm{v})\right.\right\}
\end{align}
where $\mathcal{P}^n$ is all probability distribution defined on the set of states $\{0, 1\}^n$.

The optimal machine can't be obtained by solving Eq. \eqref{fun:loss_function} directly. However, the suboptimal solution to the training problem can be found using the method of alternating minimization\cite{csisz1984information,amari1992information,byrne1992alternating}, which is a powerful method to minimize known loss function between two sets. The loss function is Kullback-Leibler divergence, and two sets are $\mathcal{B}$ and $\mathcal{E}_{Q(V)}$ in Boltzmann machine learning problem. A sequence of distributions $\{P_1,Q_1,P_2,Q_2,\cdots\}$ will be yielded when training Boltzmann machine through alternating minimization and these distributions satisfy
\begin{align}
	\label{fun:alm_first}
	Q_t(V, H) &= \min_{q \in \mathcal{E}_{Q(V)}}D(q\|P(V,H;\theta_t))\\
	\label{fun:alm_second}
	P(V, H;\theta_{t+1}) &= \min_{p \in \mathcal{B}} D(Q_t(V,H)\|p)	
\end{align}

The key point of alternating minimization in Boltzmann machine is how to solve Eq. \eqref{fun:alm_first} and Eq. \eqref{fun:alm_second}. Each iteration of alternating minimization is divided into two stages. At time $t$, the parameters of Boltzmann machine are fixed, and the best $Q_{t}(V,H)$ in $\mathcal{E}_{Q(V)}$ can be obtained through solving Eq. \eqref{fun:alm_first} in the first stage. In this stage, both Byrne and Amari et al. stated that solution of Eq. \eqref{fun:alm_first} had the following form:
\begin{align}
    \label{fun:alm_first_solution}
	q_{t}(\bm{v},\bm{h}) = q(\bm{v})p(\bm{h}|\bm{v}; \theta_{t})
\end{align}
Then new machine that is closest to $Q_{t}(V, H)$ can be obtained through solving Eq. \eqref{fun:alm_second} in the second stage.
The learning problem there is to construct $n$-unit Boltzmann machine to approximate $Q_{t}(V, H)$. Byrne solved this problem through Iterative Proportional Fitting, and Amari et al. through $m$-geodesic projection.

\subsection{Monte Carlo EM in Boltzmann Machine}
The Boltzmann machine can be trained through MCEM in theory. Firstly, based on the known visible states, we can draw hidden states from $p(\bm{h}|\bm{v})$ in the E-step. Secondly, new parameters can be given in the M-step through maximizing 
\begin{align}
\label{fun:bm_mcem}
\frac{1}{K}\sum_{k=1}^K \log p(\bm{v}^{(k)}, \bm{h}^{(k)};\theta)
\end{align}
A general approach to maximize this function is gradient descent, and update rules are
\begin{align}
	\label{fun:bm_mcem_update}
	\Delta w_{ij} = \langle x_ix_j \rangle_{q_t} - \langle x_ix_j \rangle_{p(x;\theta)}
\end{align}
where $\{\bm{x}^{(k)}\} = \{\bm{v}^{(k)}, \bm{h}^{(k)}\}$. Direct computation of average with respect to $p(x;\theta)$ is difficult. This is major bottleneck in training Boltzmann machine.

\section{Learning Boltzmann Machine With EM-Like Method}
We have note that $\langle x_ix_j \rangle_{p(x;\theta)}$ can't calculated directly. In fact, the maximization in the M-step of MCEM is finding a generative distribution behind complete-data set $\{\bm{x}^{(k)}\} = \{\bm{v}^{(k)}, \bm{h}^{(k)}\}$ with fully visible Boltzmann machine. Based on this idea, we propose EM-like method. In the E-step, we use a Monte Calo approximation for the intractable expectation. In the M-step, we replace the intractable likelihood objective with other objective.

Based on this idea, the intractable likelihood objective can be replaced by pseudo-likelihood, and this method is called pseudo-EM\cite{liang2003maximum,xiang2008pseudolikelihood}. Of course, this objective can be also replaced by composite likellihood\cite{asuncion2010learning,varin2011overview}.

The common method to maximize above objective function is gradient descent. Actually, we can only approximate $\Delta w_{ij}$ in the M-step of EM-like method. The detail is exhibited in Alg.\ref{alg:bm-learning}.

To resolve this approximation, we can adopt a distribution $r(x;\theta)$ to approxiamting $p(x;\theta)$ in Eq.\eqref{fun:bm_mcem_update}. Update rules here are
\begin{align}
	\Delta w_{ij} =\langle x_ix_j \rangle_{q_t} - \langle x_ix_j \rangle_{r(x;\theta)}
\end{align}
Contrastive divergence\cite{hinton2002training,carreira2005contrastive} and its modification, persistent contrastive divergence\cite{tielemanPcd2008}, are based on such an idea.

\begin{algorithm}[hbt]
\caption{Learning Boltzmann machine with EM-like method}
\label{alg:bm-learning}
\begin{algorithmic}[1]
\STATE Given: a training set of $K$ vectors $\{\bm{v}^{(1)}, \cdots, \bm{v}^{(K)}\}$.
\STATE Randomly initialize parameters $\theta_0$ of Boltzmann machine and let $t = 0$.
\WHILE{The termination condition can't be satisfied}
	\FOR{$k = 1$ to $K$}
		\STATE $\bm{h}_t^{(k)} \sim p(\bm{h}|\bm{v}^{(k)}, \theta_t)$
	\ENDFOR
	\STATE We subdivide complete-data set $\{\bm{v}^{(k)}, \bm{h}_t^{(k)}\}$ into M mini-batches and let $W_{t+1,0} = W_{t, M}$, $b_{t+1,0} = b_{t,M}$.
	\FOR{$j = 1$ to $M$}
	\STATE Based on current mini-batch, $\Delta W$ and $\Delta b$ can be computed by contrastive divergence, persistent contrastive divergence, pseuo-likelihood estimation or other methods which can train fully visible Boltzmann machine. After that, update the parameter through:
	\begin{align}
	 W_{t+1,j} &= W_{t+1,j-1} + \alpha_t \Delta W\\
	 b_{t+1,j} &= b_{t+1,j-1} + \alpha_t \Delta b
	\end{align}
	\ENDFOR
   \STATE Decrease $\alpha_t$ and let $\theta_{t+1} = \{W_{t+1,M}, b_{t+1, M}\}$.
	\STATE Let $t = t + 1$.
\ENDWHILE
\end{algorithmic}
\end{algorithm}

\subsection{Relationship of EM-like Method with Alternating Minimization}
\label{sec:eml_am}
There are three distributions in Eq. \eqref{fun:alm_first} and Eq. \eqref{fun:alm_second}: data distribution $Q(V)$, united distribution $Q_{t}(V, H)$ and model distribution $P(V, H;\theta_{t+1})$. In EM-like method, these three distributions are substituted by data set $\{\bm{v}^{(k)}\}$, complete-data set $\{\bm{v}^{(k)}, \bm{h}_t^{(k)}\}$ and parameters $\theta_{t+1}$ respectively.

$P(V, H;\theta)$ can be substituted by $\theta$, because the relationship between model distribution and parameters of Boltzmann machine is bijective\cite{amari1992information}. $\bm{v}^{(k)}$ is drawn from $q(\bm{v})$ and $\bm{h}_t^{(k)}$ from $p(\bm{h}|\bm{v}^{(k)};\theta_t)$. This suggests that $(\bm{v}^{(k)}, \bm{h}_t^{(k)})$ is drawn from $q_{t}(\bm{v}, \bm{h})$. Based on Glivenko-Cantelli theorem, unknown distribution can be substituted by data set sampling from this distribution. $Q(V)$ and $Q_{t}(V, H)$ can be substituted with data set $\{\bm{v}^{(k)}\}$ and complete-data set $\{\bm{v}^{(k)}, \bm{h}_t^{(k)}\}$ respectively. This suggests $Q_{t}(V, H)$ is gotten after E-step in EM-like method.

The learning problem addressed in the M-step of EM-like method is to find Boltzmann machine to approximate complete-data set $\{\bm{v}^{(k)}, \bm{h}_t^{(k)}\}$, when the target in the second stage of alternating minimization is to find optimal Boltzmann machine without hidden units to approximate $Q_t(V, H)$. It's worthy to note that the approximation in EM-like method may not be optimal.

\subsection{Relationship of EM-like Method with Contrastive Divergence in Restricted Boltzmann Machine}
\label{sec:eml_cd}
\subsubsection{Contrastive Divergence in Restricted Boltzmann Machine}
Contrastive divergence was firstly proposed by Hinton in 2002. After that, Hinton and Salakhutdinov\cite{hinton2006reducing} applied it to training restricted Boltzmann machine, where update rules of contrastive divergence were
\begin{align}
	\label{fun:cd_rbm_original}
	\nonumber\Delta w_{ij} = &\sum_{\bm{v}}q_0(\bm{v})\sum_{\bm{h}} p(\bm{h}|\bm{v}; \theta) v_ih_j\\
	 & - \sum_{\bm{v}}q_k(\bm{v})\sum_{\bm{h}} p(\bm{h}|\bm{v}; \theta) v_ih_j
\end{align}
where $q_k = q_0T^k$. In Hinton's sampling scheme, one step of Gibbs sampling was carried out in two half-steps. That is $T$ equals $T_vT_h$, where $T_{v;i,j} = p(\bm{v}=i|\bm{h}=j)$ and $T_{h;i,j} = p(\bm{h}=i|\bm{v}=j)$.

\subsubsection{EM-like Method in Restricted Boltzmann Machine}
We now know the relationship of EM-like method with alternating minimization and the best $Q_t(V,H)$ will be gotten through Eq.\eqref{fun:alm_first_solution} after E-step. We define
\begin{align}
q_0(\bm{v},\bm{h}) = q_0(\bm{v})p(\bm{h}|\bm{v};\theta)
\end{align}

If we adopt distribution $q_k(v,h)$ to approxiamting $p(v,h;\theta)$ in Eq.\eqref{fun:bm_mcem_update}, the update rules here will be
\begin{align}
	\label{fun:rbm_am_update}
	\Delta w_{ij} = \sum_{\bm{v},\bm{h}}q_0(\bm{v},\bm{h})v_ih_j - \sum_{\bm{v},\bm{h}}q_k(\bm{v},\bm{h})v_ih_j
\end{align}
If we get $q_k(\bm{v})$ through Hinton's sampling scheme and define
\begin{align}
q_k(\bm{v},\bm{h}) = q_k(\bm{v})p(\bm{h}|\bm{v};\theta)
\end{align}
it's obvious that Eq.\eqref{fun:cd_rbm_original} and Eq.\eqref{fun:rbm_am_update} are exactly the same thing. That is to say, contrastive divergence in restricted Boltzmann machine is only special case of EM-like method.

\section{Evaluating Boltzmann Machine}
Evaluating how the quality of Boltzmann machine meets the given data set plays an important role in model selection and model comparison. Although Kullback-Leibler divergence is the best criterion, it is not practical because of the high dimension of $Q(V)$. In this paper, a new criterion, \emph{avg-error}, is proposed.

We define $\bar{q}(\bm{x}; \theta) = q(\bm{v})p(\bm{h}|\bm{v}; \theta)$ and 
\begin{equation}
\left\{ \begin{aligned}
	q_i =& \sum_{\bm{x}} \bar{q}(\bm{x}; \theta)x_i\\
	q_{ij} =& \sum_{\bm{x}} \bar{q}(\bm{x}; \theta)x_ix_j
\end{aligned} \right. \& \left \{
\begin{aligned}
	p_i =& \sum_{\bm{x}} p(\bm{x}; \theta)x_i\\
	p_{ij} =& \sum_{\bm{x}} p(\bm{x}; \theta)x_ix_j
\end{aligned} \right.
\end{equation}
Then the quality of Boltzmann machine can be evaluated through
\begin{equation}
	\label{fun:avg-error}
	\sum_{i=1}^{m-1}\sum_{j=i+1}^{m}(p_{ij} - q_{ij})^2 + \sum_{i=1}^m(p_i - q_i)^2
\end{equation}
We name this value after \emph{avg-error}.

The gradient of the Kullback-Leibler divergence $D(Q(V)\|P(V; \theta))$ can be given by:
\begin{align}
	\label{fun:kl_partial_b}
	\frac{\partial D(Q(V)\|P(V; \theta))}{\partial b_i} =& -q_i + p_i\\
	\label{fun:kl_partial_w}
	\frac{\partial D(Q(V)\|P(V; \theta))}{\partial w_{ij}} =& -q_{ij} + p_{ij}
\end{align}
When the \emph{avg-error} equals to 0, $p_{ij}$ equals to $q_{ij}$ for all $i,j$ and the best parameters which minimize $D(Q(V)\|P(V; \theta))$ are obtained.

$\{p_i, p_{ij}| 1\le i\le n, i < j\}$ can determine the Boltzmann machine with $n$-unit. $\{q_i, q_{ij}| 1\le i\le m, i < j\}$ can represent distribution $Q(V)$ to some extent. So, \emph{avg-error} can assess the similarity of two distributions.

The validity of the \emph{avg-error} can be demonstrated by the Kullback-Leibler divergence curves and \emph{avg-error} curves in Fig.\ref{fig:little_kl} and Fig.\ref{fig:little_avg}. Kullback-Leibler divergence curves and \emph{avg-error} curves have similar trend in these two figures. Particularly, when Kullback-Leibler divergence is close to an extremum, \emph{avg-error} is also close to an extremum.

When $i \leq  m$ and $j \leq m$, $q_i$ and $q_{ij}$ can be appoximated in data set $\mathcal{D}$ with $K$ elements as 
\begin{align}
	q_{i} \approx & \frac{\sum_{k=1}^KI_{\{v^{(k)}_i = 1\}}}{K} \\
	q_{ij} \approx& \frac{\sum_{k=1}^KI_{\{v^{(k)}_i = 1, v^{(k)}_j = 1\}}}{K}
\end{align}
where $\bm{v}^{(k)} \in \mathcal{D}$. $I_{\{v^{(k)}_i = 1, v^{(k)}_j = 1\}} = 1$ if $v^{(k)}_i = 1$ and $v^{(k)}_j = 1$, then $I_{\{v^{(k)}_i = 1, v^{(k)}_j = 1\}} = 0$ otherwise.

Geman and Geman\cite{geman1984stochastic} proved that expectation of function $f(x)$ in Boltzmann machine could be approximated by the usual ergodic average. That is to say, if $\bm{x}_1, \bm{x}_2, ..., \bm{x}_R$ is drawn from Boltzmann machine with parameter $\theta$ by Gibbs sampling, expectation of $f(\bm{x})$ can be approximated by:
\begin{align}
	\sum_{\bm{x}}p(\bm{x};\theta)f(\bm{x}) \approx \frac{1}{R} \sum_{t=1}^Rf(\bm{x}_t)
\end{align}

Only if $i \neq j$, Boltzmann machine has the following property:
\begin{align}
	\sum_{\bm{x}}p(\bm{x};\theta)x_i =& \sum_{\bm{x}}p(\bm{x};\theta)p(x_i=1|\bm{x}_{-i};\theta)\\
	\sum_{\bm{x}}p(\bm{x};\theta)x_ix_j =& \sum_{\bm{x}}p(\bm{x};\theta)p(x_i=1|\bm{x}_{-i};\theta)x_j
\end{align}
Then, $p_i$ and $p_{ij}$ can be approximated by
\begin{align}
	p_{i} \approx & \frac{\sum_{k=1}^Rp(x_i = 1|\bm{x}^{(k)}_{-i};\theta)}{R} \\
	p_{ij} \approx & \frac{\sum_{k=1}^Rp(x_i = 1|\bm{x}^{(k)}_{-i};\theta)x^{(k)}_j}{R}
\end{align}
where $\bm{x}$ is drawn from Boltzmann machine with parameter $\theta$ by Gibbs sampling; $R$ is number of samples.

The sampling time complexity is $O(Rn)$ in this evaluating method. Computational complexity of all $p(x_i|\bm{x}_{-i}^{(k)}; \theta)$ is $O(Rmn)$ rather than $O(Rn^2)$, because $i \leq m$.  There are total $\frac{m(m+1)}{2}$ terms in Eq. \eqref{fun:avg-error}, and computational time of approximating each term is about $O(R)$. Therefore, the total computational complexity of evaluating method is $O(Rmn)$.

\section{Experimental Results}
An artificial data set and MNIST\cite{mnist_url} are used in the experiments. To speed up learning, data set is subdivided into mini-batches, and the parameters are updated after each mini-batch. The learning rate is fixed at $0.007$. To evaluate learning parameters, Kullback-Leibler divergence of data distribution with model distribution or \emph{avg-error} of data set with Boltzmann machine is computed after each epoch.

If the units of Boltzmann machine are less than 30, both Kullback-Leibler divergence and \emph{avg-error} will be computed, then only \emph{avg-error} be computed, otherwise. To approximate $p_{ij}$ more accurately, $R$ equals $1000n$ in the experiments.

\subsection{Data Sets}
Each element in artificial data set is drawn independently from a given distribution with $2^{13}$-dimension. We learn Boltzmann machine to approximate this set, which has $13$ visible units and $7$ hidden units.

The MNIST digit data set contains $60000$ training and $10000$ testing images, and each image is $28 \times 28$ pixels. We all know that each pixel in MNIST ranges from $0$ to $255$. To adapt to our model, we firstly convert each image to binary by the threshold of $128$. This set is approximated by $1000$-unit Boltzmann machine.

\subsection{Experiments in Restricted Boltzmann Machine}
Sampling is relatively simple in restricted Boltzmann machine, because there are no intralayer connections. Particularly, when the visible states are known, the hidden states can be directly drawn, and vice versa. We show the performance of contrastive divergence, persistent contrastive divergence and EM-like method. Contrastive divergence, persistent contrastive divergence and pseuo-likelihood estimation are used in the M-step of EM-like method.

\begin{figure}[htb]
  \centering
  \subfloat[k = 1, batchsize = 500]{
    \includegraphics[width=0.37\textwidth]{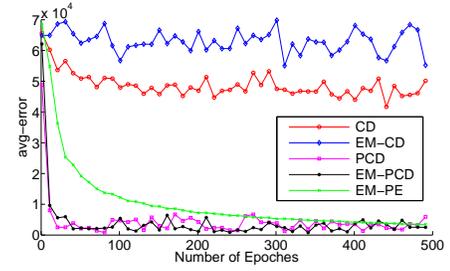}
    \label{fig:rbm-mnist-1-500}
  }\\
  \subfloat[k = 1, batchsize = 5000]{
    \includegraphics[width=0.37\textwidth]{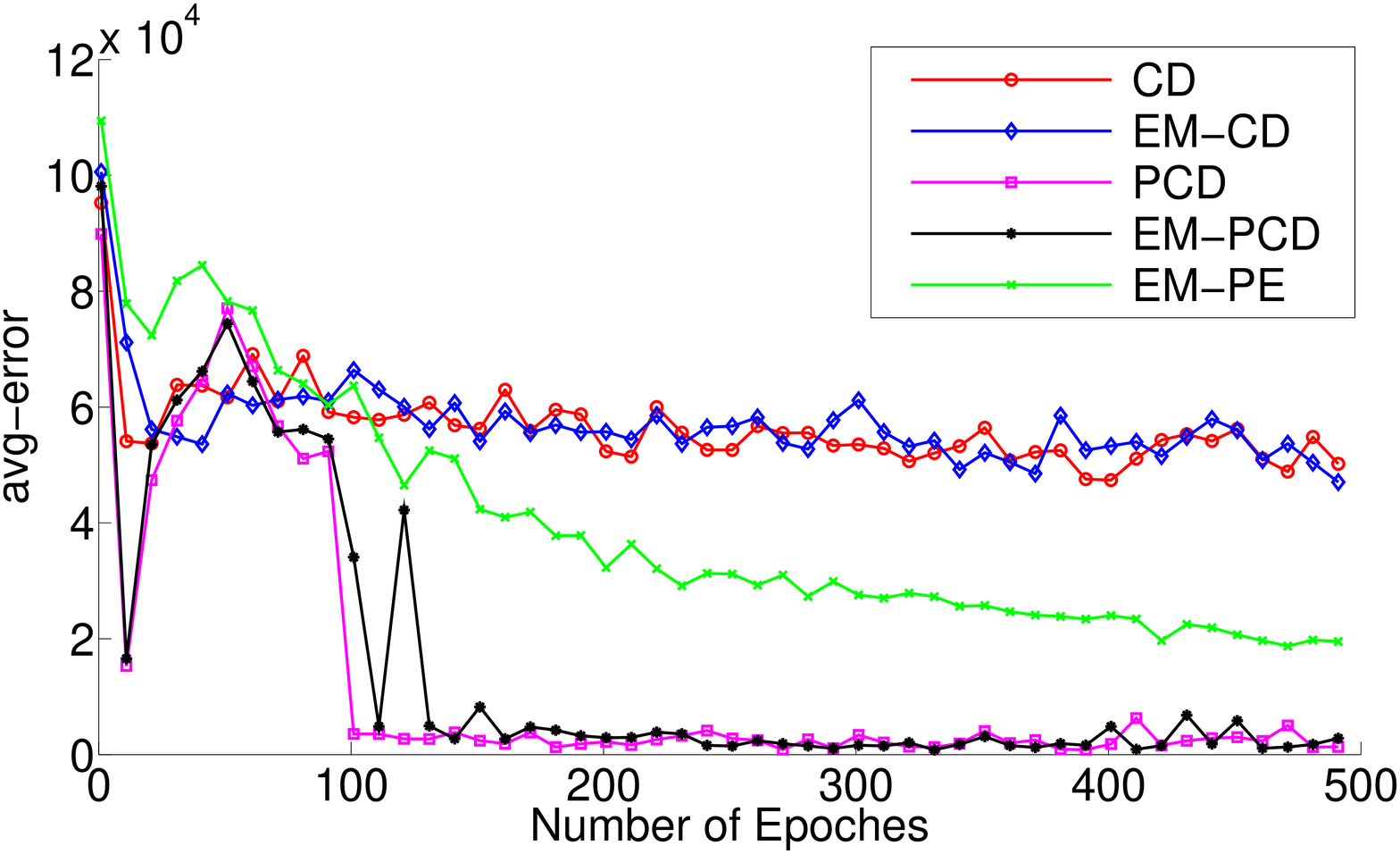}
    \label{fig:rbm-mnist-1-5000} 
  }\\
  \subfloat[k = 10, batchsize = 500]{
    \includegraphics[width=0.37\textwidth]{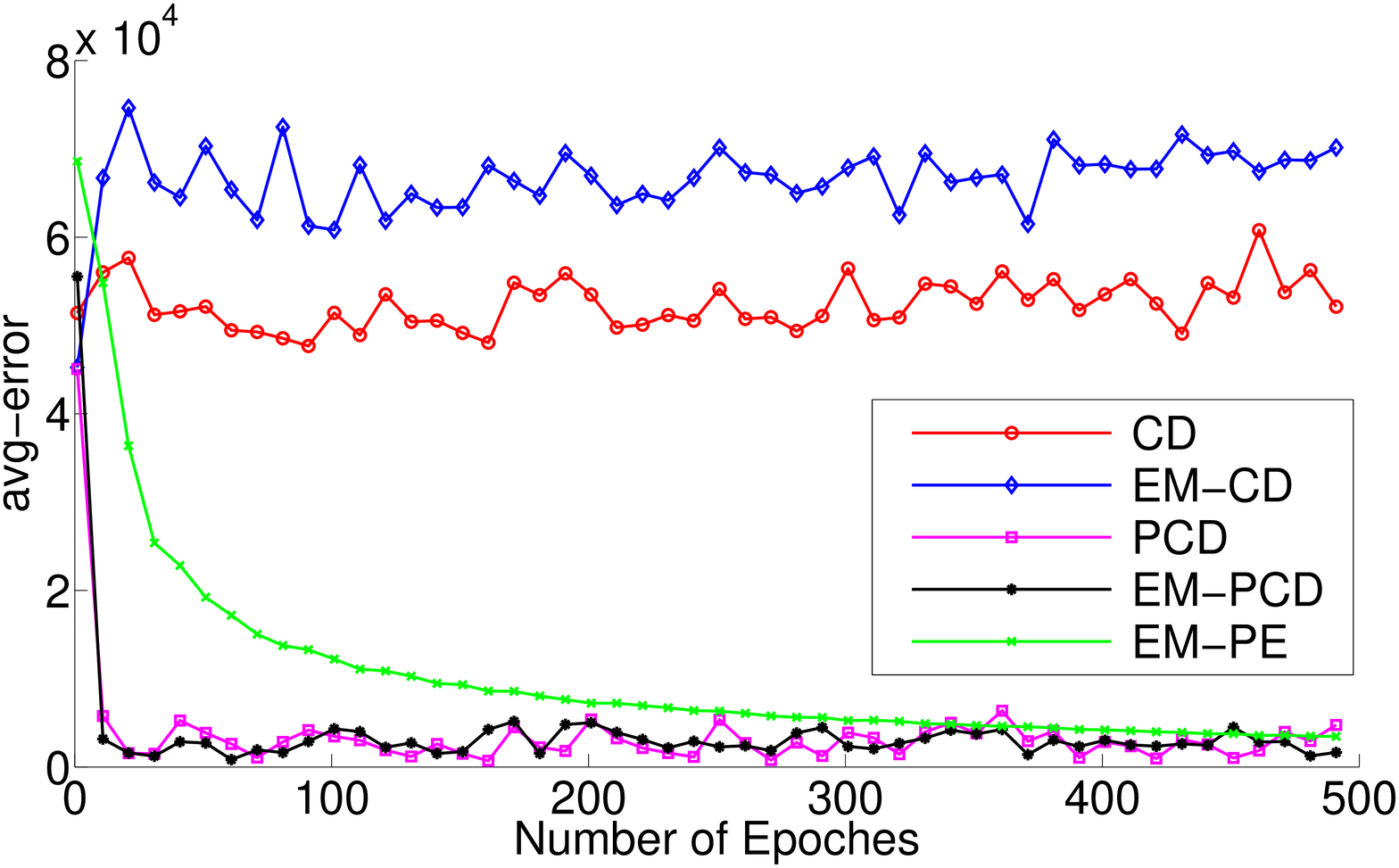}
    \label{fig:rbm-mnist-10-500} 
  }\\
  \subfloat[k = 10, batchsize = 5000]{
    \includegraphics[width=0.37\textwidth]{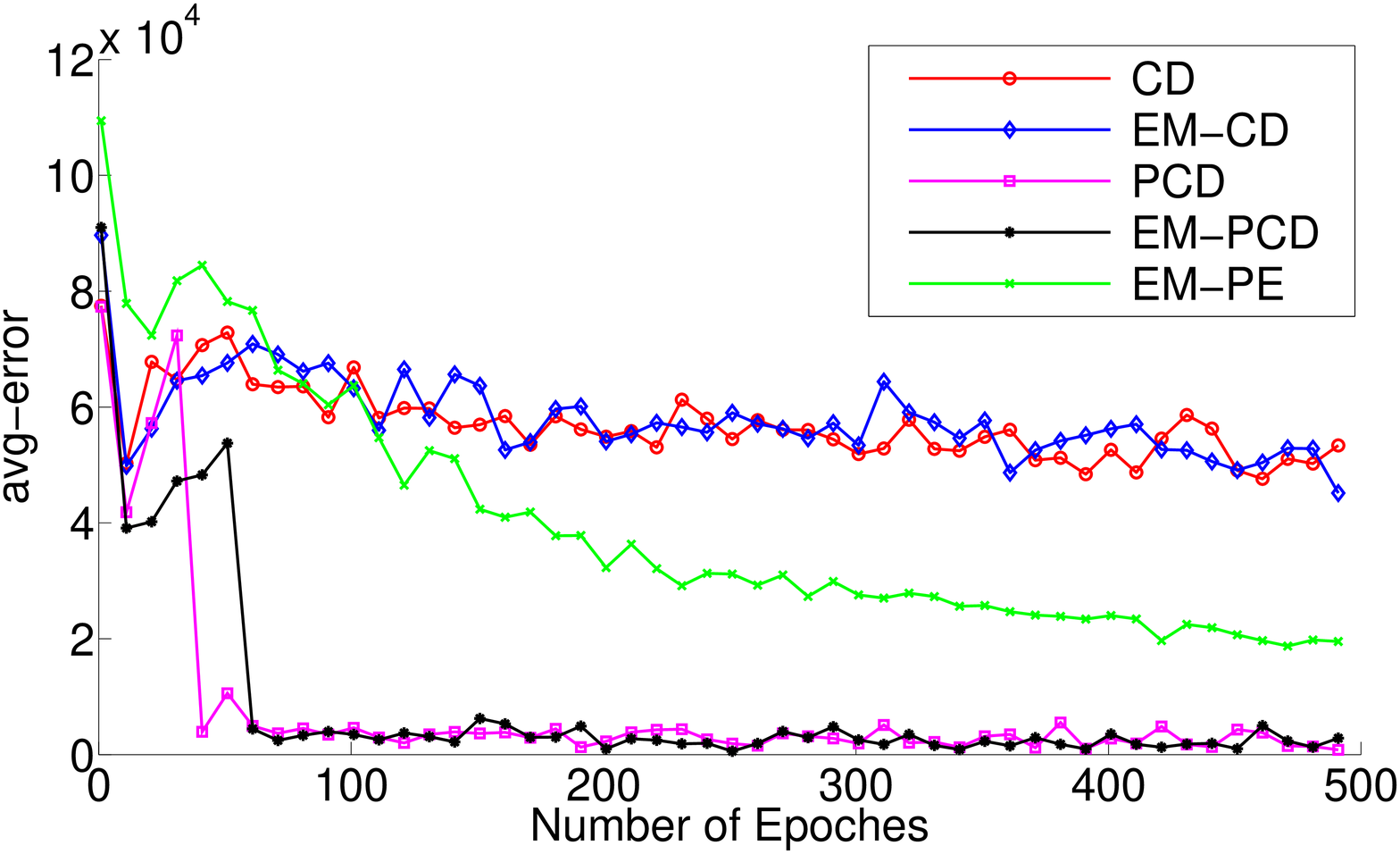}
    \label{fig:rbm-mnist-10-5000} 
  }
  \caption{Performance of different methods in restricted Boltzmann machine, where $k$ is steps of Gibbs sampling used in contrastive divergence and persisent contrastive divergence and each mini-batch has `batchsize' cases.}
  \label{fig:rbm-mnist} 
\end{figure}

The result of training Boltzmann machine can refer to Fig.\ref{fig:rbm-mnist} and Tab.\ref{tab:ais}. When hidden states of entire data set are drawn at the start of outer iteration in EM-like method, hidden states of only mini-batch are drawn  immediately after parameter updating in Hinton's method. That's why there is a gap between CD curves and EM-CD curves in Fig.\ref{fig:rbm-mnist-1-500} and Fig.\ref{fig:rbm-mnist-10-500}. From Fig.\ref{fig:rbm-mnist}, when steps of Gibbs sampling($k$) is larger, this gap disappears. At the same time, curves generated by persistent contrastive divergence and EM-PCD overlap. These all can prove the discussion in Sec.\ref{sec:eml_cd} that contrastive divergence used in restricted Boltzmann machine by Hinton is the special case of EM-like method. 

We can obtain a estimate of $Z$ in RBM through AIS\cite{neal2001annealed}. The estimate of $Z$ and the log probability of test data can refer to Tab.\ref{tab:ais}. These results are computed after training. Broadly speaking, the lower \emph{avg-error}, the lower likelihood. This proves again that our \emph{avg-error} is practicable.

\begin{table*}[htb]
  \centering
  \caption{Results of estimating partition functions of restricted Boltzmann machine, the estimates of the average training and test log probabilities, the estimates of the training and test \emph{avg-error}.}
  \label{tab:ais}
  \begin{tabular}{p{2cm}ccccc}
	  \hline
	  \multirow{2}{*}{} & \multirow{2}{*}{$\log \hat{Z}$} & \multicolumn{2}{c}{Average Log Probability}&\multicolumn{2}{c}{\emph{avg-error}}\\
		&  & Train & Test & Train & Test\\
	\hline
	EM-PE& 481.60 & -320.54 & -319.17 & 21254.72 & 21010.64\\
	\hline
	CD& 777.37 & -503.61 & -502.20 & 51788.62 & 51043.02\\
	\hline
	EM-CD& 784.57 & -503.44 & -502.05 & 52962.77 & 52236.90\\
	\hline
	PCD & 652.35 & -139.78 & -138.26 & 1655.68 & 2237.39\\
	\hline
	EM-PCD& 658.21 & -139.89 & -138.55 & 765.11 & 1313.37\\
	\hline
  \end{tabular}
\end{table*}
\subsection{Experiments in Boltzmann Machine}
\begin{figure}[htb]
  \centering
  \subfloat[Contrastive divergence]{
    \includegraphics[width=0.37\textwidth]{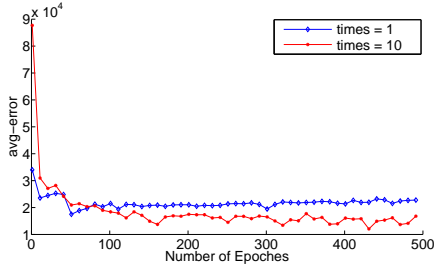}
    \label{fig:cd_gr}
  }\\
  \subfloat[Persistent constrastive divergence]{
    \includegraphics[width=0.37\textwidth]{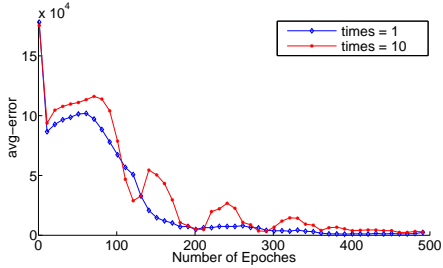}
    \label{fig:pcd_gr} 
  }\\
  \subfloat[Pseuo-likelihood estimation]{
    \includegraphics[width=0.37\textwidth]{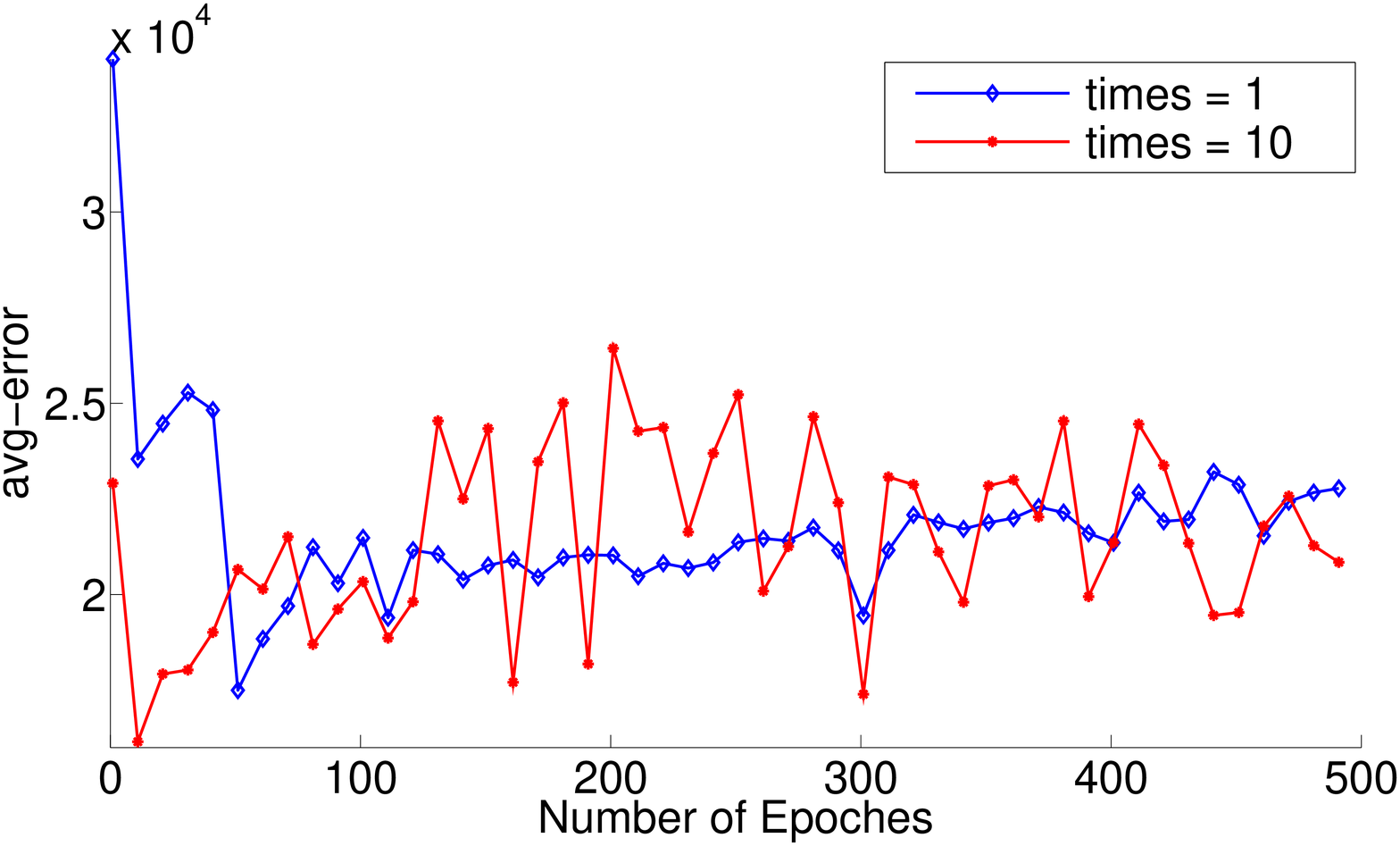}
    \label{fig:pe_gr} 
  }
  \caption{Performace under $n - m$ and $10(n - m)$ Gibbs sampling in the E-step of EM-like method. Contrastive divergence, persistent contrastive divergence and pseuo-likelihood estimation are used in the M-step. Each mini-batch has $500$ cases.}
  \label{fig:gibbs_round} 
\end{figure}

Compared with restricted Boltzmann machine, the hidden states in Boltzmann machine with any pattern of connectivity are more diffcult to draw. Gibbs sampling is adopted in our experiment. Firstly, a unit $r$ is chosen randomly in all hidden units. Secondly, new state of this unit is drawn from
\begin{align}
	p(h_r|\bm{v}, \bm{h}_{-r}; \theta)
\end{align}
In order to reach the stationary distribution, the two-step procedure needs to run many times and that can be very expensive.We show the influence of different steps of Gibbs sampling in Fig.\ref{fig:gibbs_round}. Although two curves of each figure in Fig.\ref{fig:gibbs_round} don't overlap, the gap is small. Especially, the gap disappears in Fig.\ref{fig:pcd_gr}. This suggests that we can only run $n-m$(the number of hidden units) Gibbs sampling in the E-step of EM-like method.

We show the performance of EM-like method in Boltzmann machine. As discussed above, $n-m$ Gibbs sampling is run in the E-step, and contrastive divergence, persistent contrastive divergence and pseuo-likelihood estimation are used in the M-step. We include varitional approximation\cite{salakhutdinov2012efficient} in the comparison. 

\begin{figure}[htb]
  \centering
  \subfloat[$D(Q(V)\|P(V;\theta_t))$]{
    \includegraphics[width=0.37\textwidth]{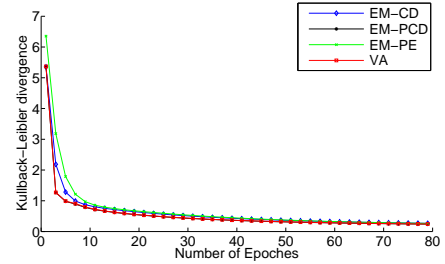}
    \label{fig:little_kl}
  }\\
  \subfloat[$D(Q_t(V, H)\|P(V, H;\theta_{t+1}))$]{
    \includegraphics[width=0.37\textwidth]{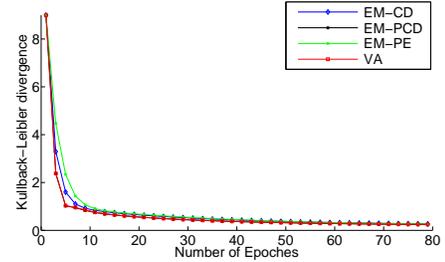}
    \label{fig:little_kl_aug} 
  }\\
  \subfloat[\emph{avg-error}]{
    \includegraphics[width=0.37\textwidth]{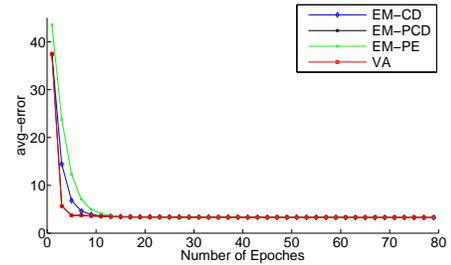}
    \label{fig:little_avg} 
  }
  \caption{Performance of different methods in Boltzmann machine
with any pattern of connectivity. This network only has 20 units. VA is short for variational approximation.}
  \label{fig:little_evaluate} 
\end{figure}

Firstly, we learn a little network to approximate artificial data set. Kullback-Leibler divergence can be computed directly in this little network. Varitional approximation and EM-like method show similar performance from Fig.\ref{fig:little_evaluate}.

The percentage in Tab.\ref{tab:difference} is computed by
\begin{align}
	\label{fun:percentage}
	\left(\sum_{\substack{1\le i,j \le m \\ i\ne j}}I_{\{|p_{ij} - q_{ij}| \lessgtr a\}}\right)/m^2
\end{align}
where $I_{\{|p_{ij} - q_{ij}| \lessgtr a\}}$ is $1$ if $|p_{ij} - q_{ij}|$ is less or greater than const $a$, then this value is $0$ otherwise. And `avg'  is computed through
\begin{align}
	\label{fun:avg}
\left(\sum_{\substack{1\le i,j \le m \\ i\ne j}}|p_{ij} - q_{ij}| + \sum_{1\le i \le m}|p_{i} - q_{i}|\right)/m^2
\end{align}

\begin{figure}[htb]
  \centering
  \subfloat[k = 10, batchsize = 500]{
    \includegraphics[width=0.37\textwidth]{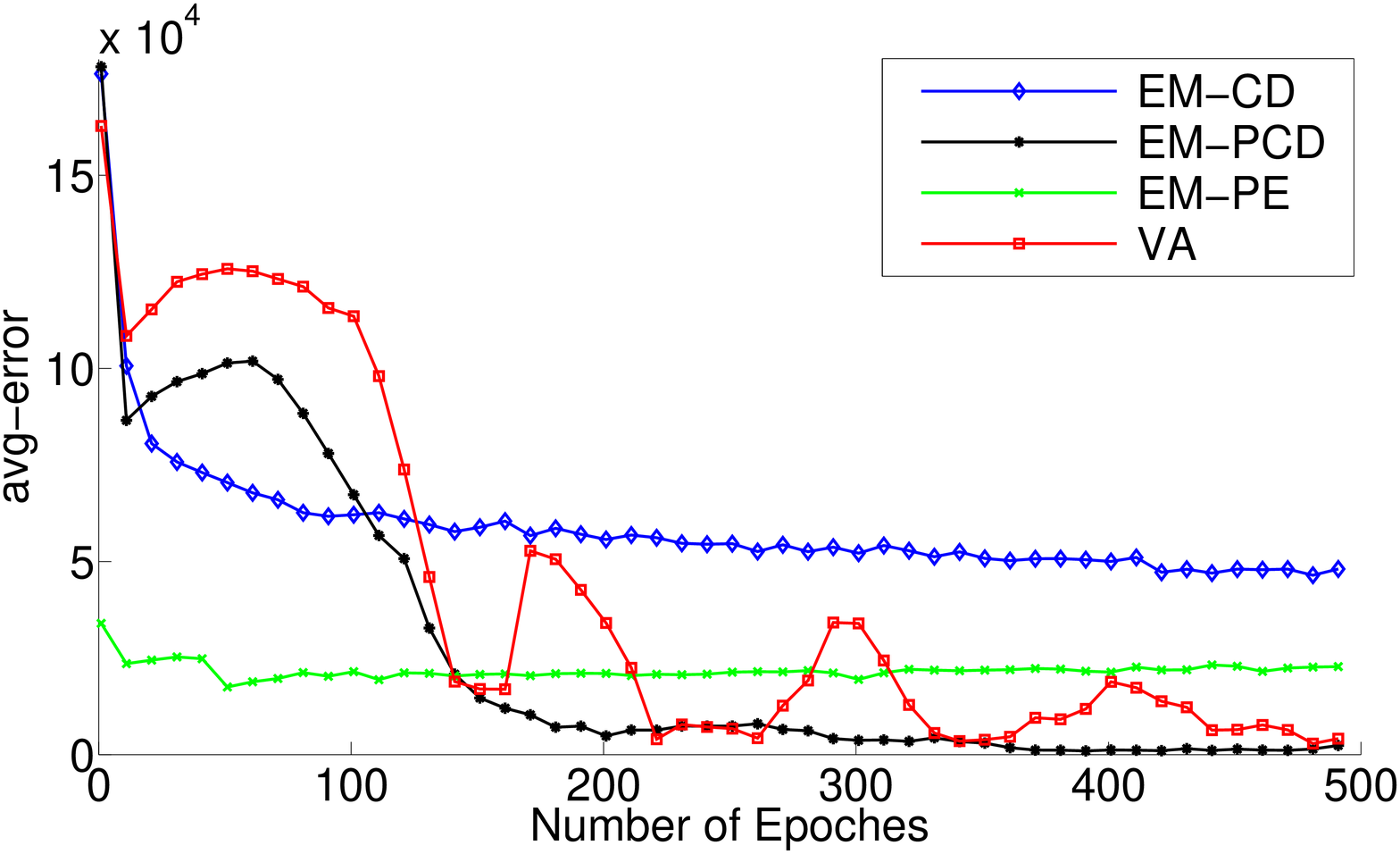}
    \label{fig:bm-mnist-10-500}
  }\\
  \subfloat[k = 10, batchsize = 5000]{
    \includegraphics[width=0.37\textwidth]{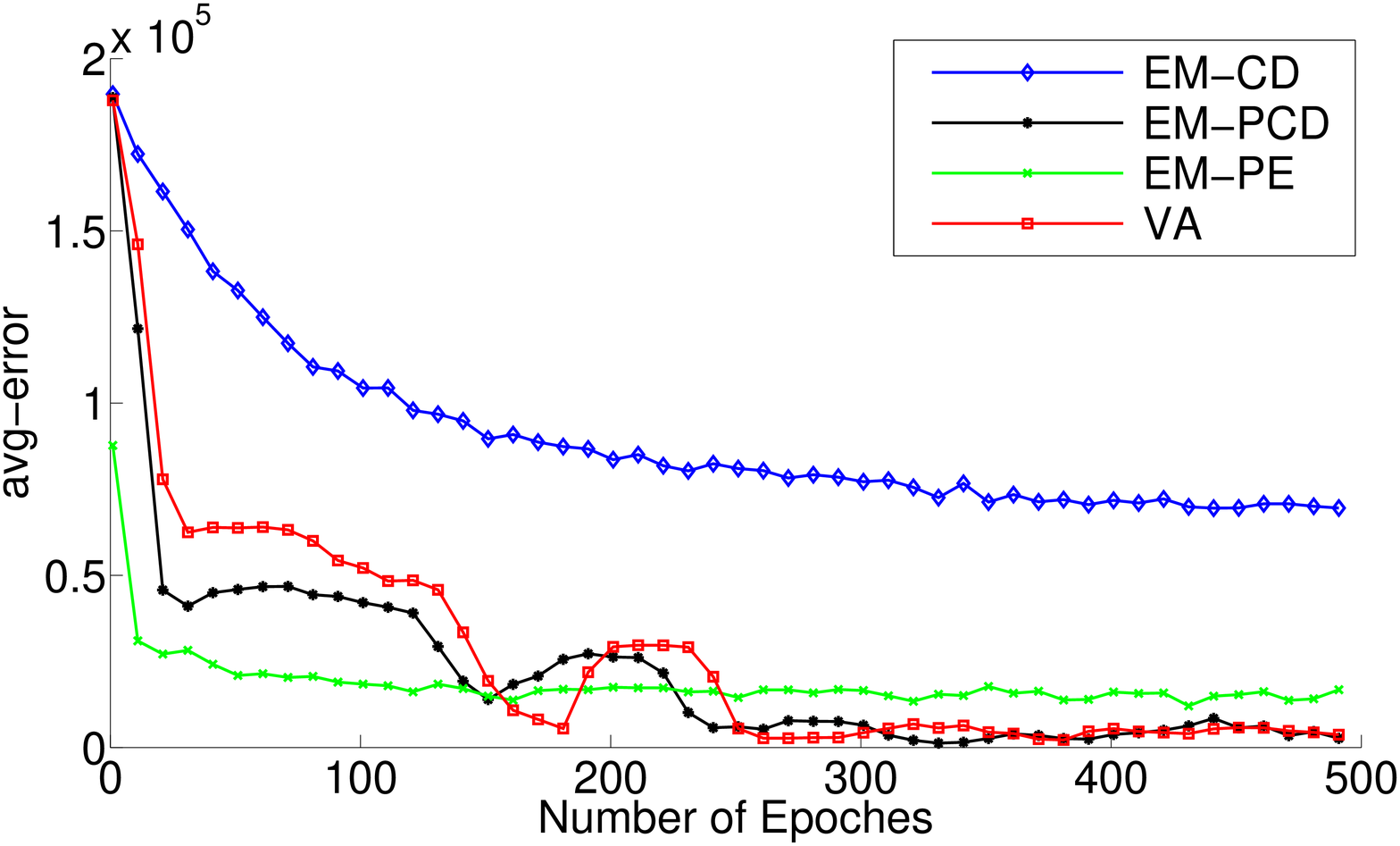}
    \label{fig:bm-mnist-10-5000} 
  }\\
  \subfloat[k = 100, batchsize = 500]{
    \includegraphics[width=0.37\textwidth]{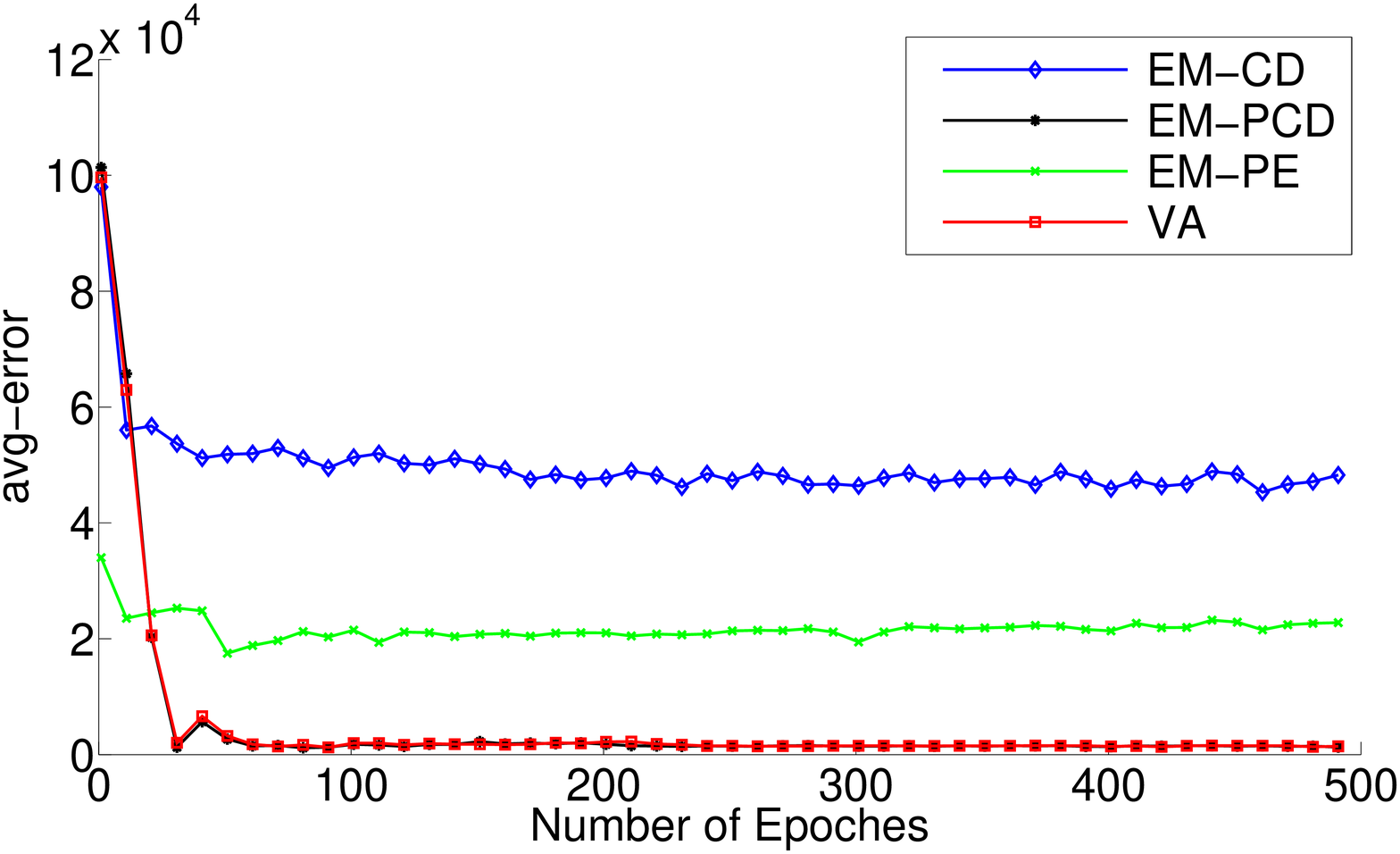}
    \label{fig:bm-mnist-100-500} 
  }\\
  \subfloat[k = 100, batchsize = 5000]{
    \includegraphics[width=0.37\textwidth]{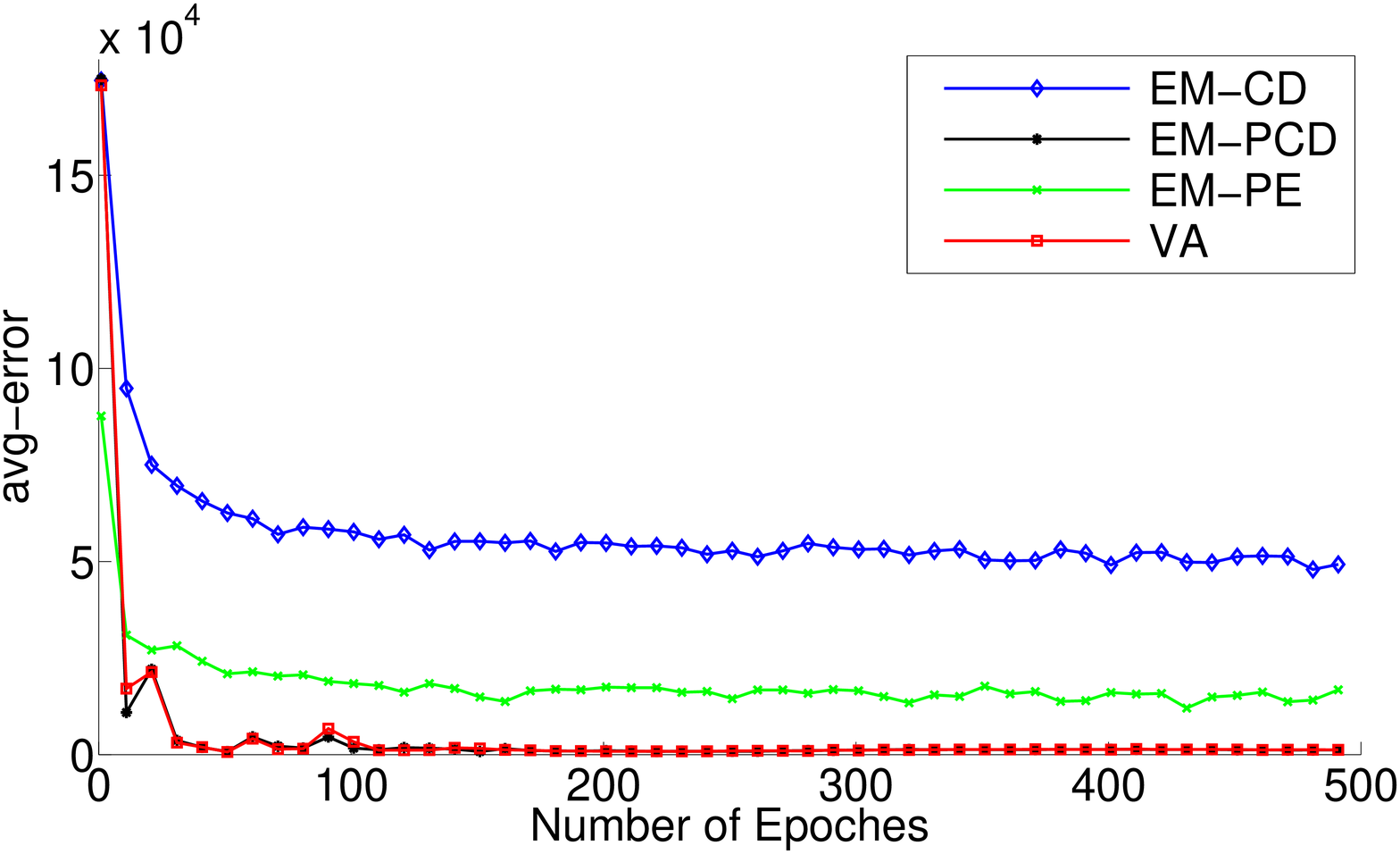}
    \label{fig:bm-mnist-100-5000} 
  }
  \caption{Performance of different methods in Boltzmann machine with any pattern of connectivity, where $k$ is steps of Gibbs sampling used in contrastive divergence and persisent contrastive divergence and each mini-batch has `batchsize' cases. VA is short for variational approximation.}
  \label{fig:bm-mnist} 
\end{figure}

As Fig.\ref{fig:bm-mnist} and Tab.\ref{tab:difference} have shown, EM-like method could train Boltzmann machine. Firstly, the curves generated by variational approximation\cite{salakhutdinov2012efficient} and EM-like method associated
with persistent contrastive divergence overlap. Secondly, the `avg' can reach to 0.02, and the \emph{avg-error} to 1245.85. The ratio of the points, where $|p_{ij} - q_{ij}|$ is less than 0.01, is close to 80\%. Compared with restricted Boltzmann machine which is trained by contrastive divergence(Hinton's method), these three values are 0.21, 53682.54 and 2\% respectively. These all suggest the model trained by EM-like method can approximate data distribution very well.

As discussed in \cite{tielemanPcd2008}, it is clear that persistent contrastive divergence outperforms the other algorithms. This result is proven again in the Tab.\ref{tab:difference}, Tab.\ref{tab:ais}, Fig.\ref{fig:rbm-mnist} and Fig.\ref{fig:bm-mnist}.

\begin{table*}[htb]
  \centering
  \caption{The results of Eq.\eqref{fun:percentage}, Eq.\eqref{fun:avg} and \emph{avg-error} after training.}
  \label{tab:difference}
  \begin{tabular}{|c|c|c|c|c|c|c|c|c|c|c|}
	\hline
	\multicolumn{2}{|c|}{}& $<0.01$ & $<0.05$ & $<0.1$ & $<0.2$ & $>0.5$ & $>0.9$ & $>0.95$ & avg & \emph{avg-error}\\
	\hline
	\multirow{5}{*}{\shortstack{R\\B\\M}} & EM-PE & 4.5\% & 30.2\% & 52.2\% & 76.3\% & 1.9\% & 0 & 0 & 0.14 & 21010.64\\
	\cline{2-11}
	& CD & 2.0\% & 14.7\% & 30.6\% & 54.7\% & 11.1\% & 0 & 0 & 0.3 & 53682.54\\
	\cline{2-11}
	& EM-CD & 1.5\% & 12.1\% & 28.4\% & 53.7\% & 10.2\% & 0 & 0 & 0.3 & 51936.81\\
	\cline{2-11}
	& PCD & 76.3\% & 89.6\% & 94.5\% & 98.4\% & 0 & 0 & 0 & 0.02 & 2905.55\\
	\cline{2-11}
	& EM-PCD & 74.1\% & 86.5\% & 92.3\% & 97.3\% & 0.2\% & 0 & 0 & 0.02 & 2147.93\\
	\hline
	\multirow{3}{*}{\shortstack{B\\M}}& EM-PE & 51.4\% & 71.4\% & 80.9\% & 90.3\% & 3.5\% & 0.3\% & 0.1\% & 0.07 & 16321.26\\
	\cline{2-11}
	& EM-CD& 24.8\% & 48.1\% & 59.8\% & 72.7\% & 12.0\% & 0.3\% & 0 & 0.17 & 48185.09\\
	\cline{2-11}
	& EM-PCD& 78.0\% & 89.3\% & 94.0\% & 98.9\% & 0 & 0 & 0 & 0.02 & 1245.85\\
	\cline{2-11}
	& VA & 77.7\% & 89.0\% & 94.2\% & 98.0\% & 0 & 0 & 0 & 0.02 & 2415.27\\
	\hline
  \end{tabular}
\end{table*}

\section{Conclusions and Future Works}
We can find out that:
\begin{enumerate}
\item The EM-like method proposed by us can train Boltzmann machine with any pattern of structure. Based on this method, contrastive divergence, persistent contrastive divergence and other methods can be applied to more complex model with latent variables.
\item 
Contrastive divergence is only special case of EM-like method in restricted Boltzmann machine.
\item $n - m$ Gibbs sampling is enough in the E-step of EM-like method. When training time is short, bigger step of Gibbs sampling used in contrastive divergence and bigger batchsize are better. When training time increases, this advantage will vanish.
\item We propose a new criterion, \emph{avg-error}, which can evaluate how the quality of Boltzmann machine meets the given data set. Its computational time is $O(Rmn)$, and it can be applied to practical applications.
\end{enumerate}

Only three methods, contrastive divergence, persistent contrastive divergence and pseuo-likelihood estimation are adopted in the M-step of EM-like method
in our experiments. Other methods which can train fully visible
Boltzmann machine will be our first work in the future.

The relationship of EM-like method with alternating minimization is discussed in Sec.\ref{sec:eml_am}. When we train Boltzmann machine through alternating minimization, the goal in the second stage is finding best machine to approximate $Q_t(V,H)$, that is
\begin{align}
	\theta_{t+1}^* = \arg \min_\theta D(Q_t(V,H)\|P(V,H;\theta))
\end{align}
Actually, this condition can be relaxed, that is, we can find $\theta_{t+1}$ which satisifies
\begin{align}
	D(Q_t(X)\|P(X;\theta_{t+1})) \leq D(Q_t(X)\|P(X;\theta_{t}))
\end{align}
where $X=\{V,H\}$.

The curves of $D(Q_t(V, H)\|P(V, H;\theta_{t+1}))$ are shown in Fig. \ref{fig:little_evaluate}. Although the fixed learnrate is used in this experiment, we can't observe that this value goes up evenly after 1000 epoches. Maybe this is the main reason why contrasive divergence converges. Convergence about contrastive divergence from this view will be our sceond work in the future.

\section{Acknowledgments}
This work is supported by the National High Technology Research and Development Plan (863 Plan) under Grant No.2011AA01A205, the National Significant Science and Technology Projects of China under Grant No.2013ZX01039001-002-003, the NSFC project under Grant No.U1433112 and No.61170253.

\bibliographystyle{IEEEtran}
\bibliography{IEEEfull,bm-ref}
\end{document}